\documentclass[conference]{IEEEtran}
\IEEEoverridecommandlockouts
\usepackage{cite}
\usepackage{amsmath,amssymb,amsfonts}
\usepackage{algorithmic}
\usepackage{graphicx}
\usepackage{textcomp}
\usepackage{xcolor}
\def\BibTeX{{\rm B\kern-.05em{\sc i\kern-.025em b}\kern-.08em
    T\kern-.1667em\lower.7ex\hbox{E}\kern-.125emX}}

\usepackage{graphicx}
\usepackage{amsmath}

\usepackage{amssymb}
\usepackage[misc,geometry]{ifsym} 
\usepackage{array}
\newcolumntype{C}[1]{>{\centering\arraybackslash}p{#1}}
\usepackage{placeins}

\usepackage{boldline, multirow}
\usepackage{cite}
\usepackage{dblfloatfix}
\usepackage{balance}
\begin{document}

\title{Leveraging Disease Progression Learning for Medical Image Recognition\\
}

\author{\IEEEauthorblockN{Qicheng Lao}
\IEEEauthorblockA{\textit{Department of Computer Science}\\ \textit{and Software Engineering} \\
\textit{Concordia University}\\
Montr\'{e}al, Qu\'{e}bec, Canada \\
qi\_lao@encs.concordia.ca}
\and
\IEEEauthorblockN{Thomas Fevens}
\IEEEauthorblockA{\textit{Department of Computer Science}\\ \textit{and Software Engineering} \\
\textit{Concordia University}\\
Montr\'{e}al, Qu\'{e}bec, Canada \\
fevens@encs.concordia.ca}
\and
\IEEEauthorblockN{Boyu Wang}
\IEEEauthorblockA{\textit{Princeton Neuroscience Institute} \\
\textit{Princeton University}\\
Princeton, New Jersey, USA \\
boyuw@princeton.edu}

}

\maketitle

\begin{abstract}
Unlike natural images, medical images often have intrinsic characteristics that can be leveraged for neural network learning. For example, images that belong to different stages of a disease may continuously follow a certain progression pattern. In this paper, we propose a novel method that leverages disease progression learning for medical image recognition. In our method, sequences of images ordered by disease stages are learned by a neural network that consists of a shared vision model for feature extraction and a long short-term memory network for the learning of stage sequences. Auxiliary vision outputs are also included to capture stage features that tend to be discrete along the disease progression. Our proposed method is evaluated on a public diabetic retinopathy dataset, and achieves about 3.3$\%$ improvement in disease staging accuracy, compared to the baseline method that does not use disease progression learning.
\end{abstract}


\section{Introduction}
Deep learning has been widely applied to medical image recognition since its great success in natural image recognition, and has achieved state-of-the-art performance in various areas such as anatomical structure identification, lesion detection and classification \cite{litjens2017survey, shen2017deep}. Unlike general object classification, medical images often have intrinsic characteristics that can be exploited to facilitate neural network learning for improved results. For example, medical recognition tasks on images acquired from computed tomography (CT) or magnetic resonance imaging (MRI) generally favor a 3D convolutional neural network (CNN) over a 2D CNN, due to the additional spatial information in three dimensions \cite{payan2015predicting, huang2017lung}. Another example of neural networks that exploit medical intrinsic information is the BrainNetCNN, where special edge-to-edge, edge-to-node and node-to-graph convolutional filters are designed to leverage topological locality of brain networks for the prediction of neurodevelopmental outcomes \cite{kawahara2017brainnetcnn}. 

\begin{figure*}[htbp]
	\centering
	\includegraphics[scale=0.42]{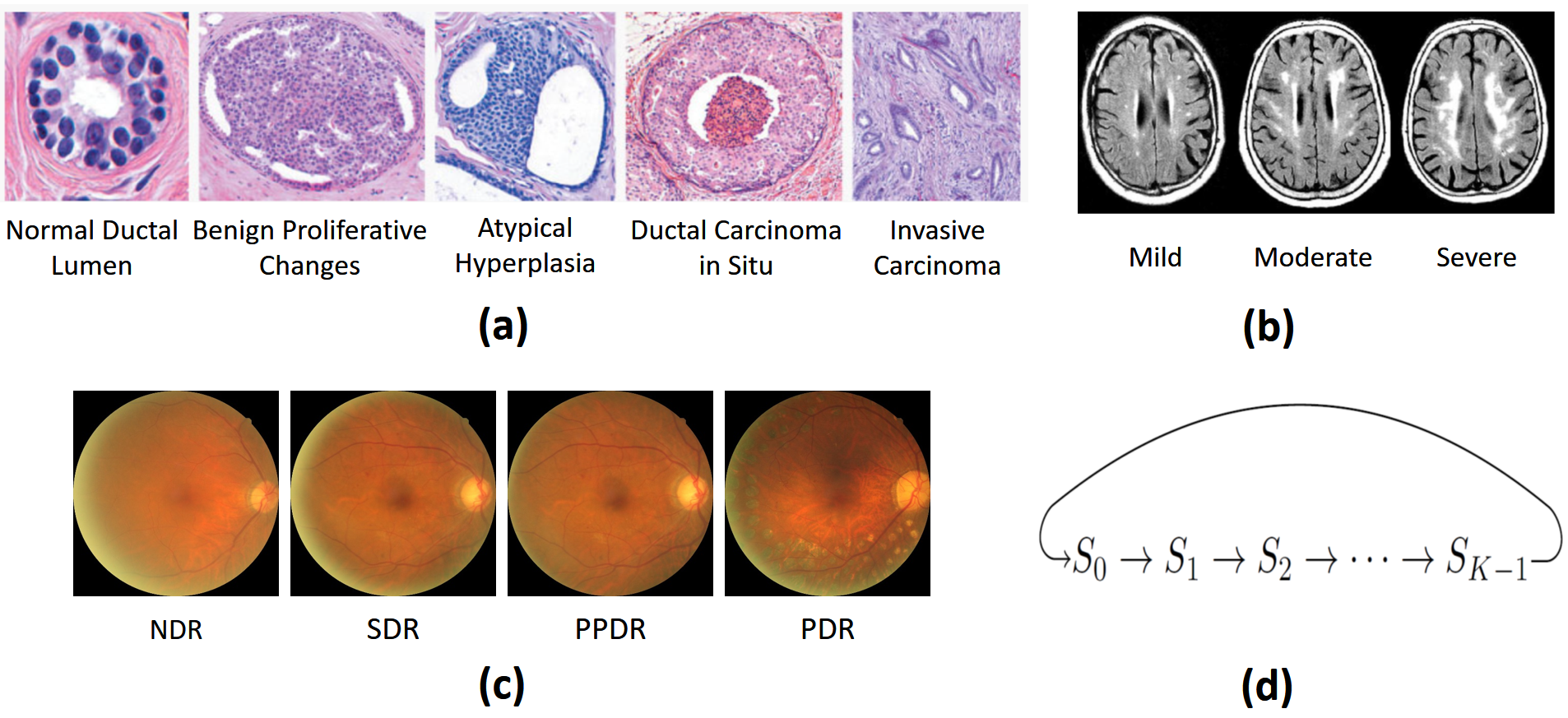}
	\caption{Examples of disease progression with different stages: (a) Breast cancer with five stages: normal, benign proliferation, atypical hyperplasia, in-situ and invasive \cite{burstein2004ductal}; (b) Aging related changes in white matter with three severity grades: mild, moderate and severe\cite{inzitari2009changes}; (c) Diabetic retinopathy with four stages: no-diabetic-retinopathy (NDR), simple-diabetic-retinopathy (SDR), pre-proliferative-diabetic-retinopathy (PPDR) and proliferative-diabetic-retinopathy (PDR) \cite{takahashi2017applying}; (d) Cyclic form of the stage sequence. }
	\label{fig_stages}
\end{figure*}

In this paper, we propose a novel method that leverages disease progression learning for medical image recognition. Concretely, given any medical recognition problem that is associated with a disease progression, we use long short-term memory (LSTM) to model the disease progression, for example, to predict survival time based on brain tumor images (e.g. short-term survival, medium-term survival and long-term survival) \cite{nie20163d}, or a disease staging problem. As illustrated in Figure \ref{fig_stages}(a, b, c), the disease staging problem commonly exists across multiple modalities, including but not limited to classifying breast histopathological images into normal, benign, in-situ or invasive \cite{burstein2004ductal}; MRI images of white matter into mild, moderate or severe based on age-related changes \cite{inzitari2009changes}; and retinal fundus images into no-diabetic-retinopathy (NDR), simple-diabetic-retinopathy (SDR), pre-proliferative-diabetic-retinopathy (PPDR) or proliferative-diabetic-retinopathy (PDR) \cite{takahashi2017applying}. Rather than considering each different stage as an independent class as done in most previous research, we hypothesize that there is a stage difference memory driven by the disease progression that should also be represented in the neural network. By leveraging this memory information from the stage sequence, more robust and optimal results could be possible.

LSTM is a widely-used network that is powerful for sequential data learning, as it has memory units that efficiently remember previous steps \cite{hochreiter1997long}. There are previous publications using LSTM for medical data, but mostly based on diagnostic text reports \cite{zhang2017mdnet}, 3D image stacks \cite{chen2016combining, xue2017full}, or clinical measurements/admissions \cite{lipton2015learning, pham2016deepcare, choi2016doctor},  among which, some also use the concept of disease progression modeling \cite{pham2016deepcare, choi2016doctor}. However, all previous work either do not clearly define stage classes for a disease progression, or still consider each stage as an independent class and the sequence learning only occurs within each individual stage class. I.e., there is no explicit learning of the stage sequence itself, other than learning of temporal sequence (e.g. a series of clinical events), or spatial sequence (e.g. 3D MRI/CT images). 

To the best of our knowledge, we are the first to use LSTM for the learning of stage sequence along the disease progression for medical image recognition, with each stage represented by feature vectors that are extracted from a well-established vision model (e.g. GoogleNet or ResNet). Auxiliary outputs from the vision model are also adopted in order to capture stage features that may not be continuous along the disease progression. Our proposed method is evaluated on a diabetic retinopathy dataset, where it shows a performance increase of around 3.3$\%$ in disease staging accuracy, compared to the baseline method that is similar but without disease progression learning.

\section{Methods}
\subsection{Disease progression learning} \label{disease_progression_learning}
Given a disease progression with $K$ sequential stages $S = \{S_k, k = 0, 1, ..., K-1\}$, with $S_{k} > S_{k-1}$ for each $k \in \left( 0,K-1 \right] $, where the greater-than sign indicates a disease progression, meaning stage $S_k$ is a subsequent stage of $S_{k-1}$, e.g. \{\textit{NDR, SDR, PPDR, PDR}\} for diabetic retinopathy and \{\textit{normal, benign, in-situ, malignant}\} for epithelial cancers, we want the neural network to learn disease stage progression by presenting the network with a sequence of images ordered by disease stage as the input $\boldsymbol{x} = [ I^{S_0}, I^{S_1}, ..., I^{S_{K-1}} ]$, where $I^{S_{k}}$ is a randomly selected image that belongs to stage $S_{k}$, and the corresponding output is simply the ordered full sequence of all disease stages: $\boldsymbol{y} = [ S_0, S_1, ..., S_{K-1} ]$. However, with the above design, all the input samples would share the same output $\boldsymbol{y}$ that is fixed by the disease stage progression, making the network training meaningless. To overcome this problem, we artificially define $S_0 > S_{K-1}$ to make the stage sequence cyclic (Figure \ref{fig_stages}(d)), so that multiple ordered full sequences of all disease stages can be generated. Therefore, $[ S_1, S_2, ..., S_{K-1}, S_0 ]$, for example, is also considered as a valid ordered full sequence containing all stages for a certain disease. As a result, the notation of stage sequence for a training sample can be updated more formally by the introduction of a modulo operation as the following:
\begin{equation*}\tag{1}\label{eq_sample_xy}
\begin{split}
\left( \boldsymbol{x}, \boldsymbol{y} \right) = \left( \right. [ I^{S_{i+0 \pmod K}}, I^{S_{i+1 \pmod K}}, ..., I^{S_{i+K-1 \pmod K}} ], \\ [ S_{i+0 \pmod K}, S_{i+1 \pmod K}, ..., S_{i+K-1 \pmod K} ] \left. \right)
\end{split}
\end{equation*}
where $i \in \left[ 0,K-1 \right]$ can be considered as the step shift size, indicating the starting stage of a sequence, and $K$ is the total number of disease stages. For simplicity of notation and explanation, we use the stage sequence of $i=0$ as an illustration in the following. 

\begin{figure*}[htbp]
	\centering
	\includegraphics[scale=0.40]{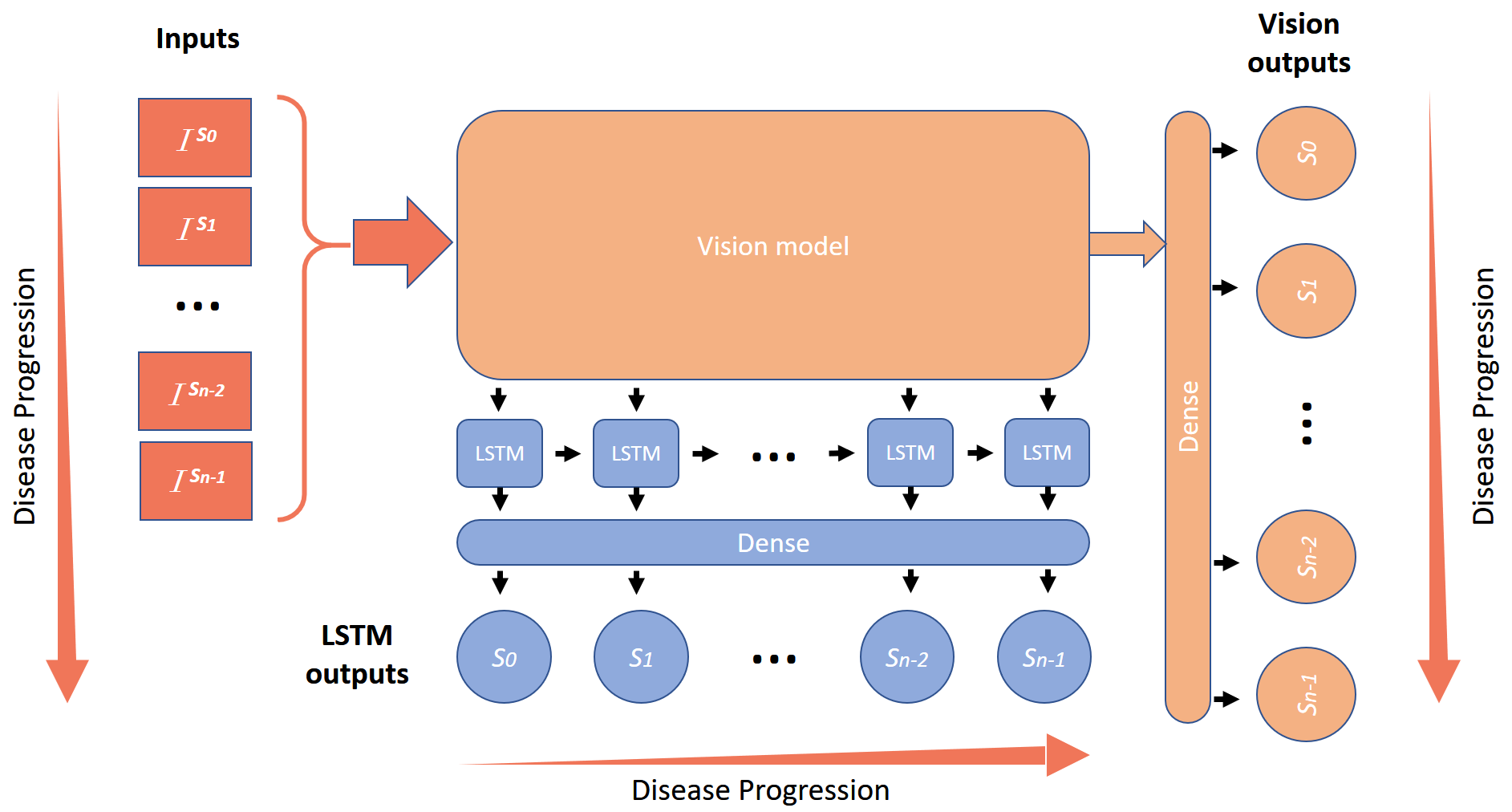}
	\caption{The architecture of our proposed network.}
	\label{fig_architecture}
\end{figure*}

As shown in Figure \ref{fig_architecture}, the proposed method contains a vision model (e.g. GoogleNet or ResNet) for the feature extraction, followed by a LSTM model for the purpose of disease progression learning. The vision model extracts a feature vector $\boldsymbol{z}^{S_{k}} \in \mathbb{R}^{C}$ from its corresponding input image $I^{S_{k}}$ for each disease stage $S_{k}$, and then the concatenated feature vector sequence of all stages $\boldsymbol{z} = [ \boldsymbol{z}^{S_{0}}, \boldsymbol{z}^{S_{1}}, ..., \boldsymbol{z}^{S_{K-1}} ] \in \mathbb{R}^{K \times C}$ is given as the input to LSTM. The LSTM maintains a hidden state $\boldsymbol{h}^{S_{k}} \in \mathbb{R}^{G}$ and a cell state $\boldsymbol{c}^{S_{k}} \in \mathbb{R}^{G}$, which are updated at each stage step $k \in \left( 0,K-1 \right]$:

\begin{equation*} \tag{2}\label{eq_lstm}
\boldsymbol{h}^{S_{k}}, \boldsymbol{c}^{S_{k}} = \text{LSTM}(\boldsymbol{z}^{S_{k}}, \boldsymbol{h}^{S_{k-1}}, \boldsymbol{c}^{S_{k-1}})
\end{equation*}
The hidden state sequence of all stages $\boldsymbol{h} = [ \boldsymbol{h}^{S_{0}}, \boldsymbol{h}^{S_{1}}, ..., \boldsymbol{h}^{S_{K-1}} ] \in \mathbb{R}^{K \times G}$ is collected from the multi-output LSTM to learn a softmax classifier $\mathcal{F}$, which finally outputs the hypotheses $\hat{\boldsymbol{y}}$ of the true stage labels $\boldsymbol{y}$:
\begin{equation*} \tag{3}\label{eq_lstm_classifier}
\hat{\boldsymbol{y}} = \mathcal{F}(\boldsymbol{h}; \boldsymbol{\theta})
\end{equation*}
where $\boldsymbol{\theta}$ is the parameter for classifier $\mathcal{F}$. Note that $\hat{\boldsymbol{y}} = [ \hat{\boldsymbol{y}}^{S_{0}}, \hat{\boldsymbol{y}}^{S_{1}}, ..., \hat{\boldsymbol{y}}^{S_{K-1}} ] \in \mathbb{R}^{K \times K}$, where each sequence element represents a probability distribution over K stage labels for its corresponding stage. Also, it is worth noting that although for the simplicity of notation, we denote both $\boldsymbol{h}$ and $\hat{\boldsymbol{y}}$ as vector sequences of all stages, the softmax classifier $\mathcal{F}$ is actually applied to each individual hidden state $\boldsymbol{h}^{S_{k}}$ independently to obtain its corresponding hypothesis $\hat{\boldsymbol{y}}^{S_{k}}$. 

Our loss function is a weighted summation of cross entropy losses at all stages:
\begin{equation*} \tag{4}\label{eq_lstm_loss}
Loss(\hat{\boldsymbol{y}}, \boldsymbol{y}) = \sum_{k=0}^{K-1} \alpha_{k} \cdot l(\hat{\boldsymbol{y}}^{S_{k}}, S_{k})
\end{equation*}
where $\alpha_k$ is the loss weight for each stage output and
\begin{equation*} \tag{5}\label{eq_cross_entropy_loss}
l(\hat{\boldsymbol{y}}^{S_{k}}, S_{k}) = \sum_{j=0}^{K-1}- \log \hat{y}^{S_{k}}_j \cdot \delta(S_{k} = S_{j})
\end{equation*}
is the cross entropy loss function, with $\hat{y}^{S_k}_j$ denoting for the probability value of image $I^{S_k}$ belonging to the stage label $S_j$. 

\subsection{Auxiliary vision outputs}
In addition to disease progression learning, the network should also be able to capture stage-wise discriminative features that tend to be discrete among different stages along the disease progression, e.g. features that only appear in a certain disease stage. Given this thought, we also design auxiliary outputs directly from the vision model extracted features $\boldsymbol{z}$ with another softmax classifier $\mathcal{F}_v$ on top of the vision model:
\begin{equation*} \tag{6}\label{eq_vision_classifier}
\hat{\boldsymbol{y}}_v = \mathcal{F}_v(\boldsymbol{z}; \boldsymbol{\theta}_v)
\end{equation*}
To distinguish the two classifiers on top of LSTM model ($\mathcal{F}_l$) and vision model ($\mathcal{F}_v$), Equation (\ref{eq_lstm_classifier}) is updated as:
\begin{equation*} \tag{7}\label{eq_lstm_classifier_new}
\hat{\boldsymbol{y}}_l = \mathcal{F}_l(\boldsymbol{h}; \boldsymbol{\theta}_l)
\end{equation*}

Similarly to LSTM outputs, the loss for auxiliary vision outputs is also defined as the weighted summation of cross entropy losses at all disease stages. Taking both losses into account, the final loss function for our proposed network is:
\begin{equation*} \tag{8}\label{eq_vision_loss}
Loss(\hat{\boldsymbol{y}}_l, \hat{\boldsymbol{y}}_v, \boldsymbol{y}) = \sum_{k=0}^{K-1} \big(\alpha_{k} \cdot l(\hat{\boldsymbol{y}}_l^{S_{k}}, S_{k}) +  \beta_{k} \cdot l(\hat{\boldsymbol{y}}_v^{S_{k}}, S_{k})\big)
\end{equation*}
The loss weights $\alpha_k$ and $\beta_k$ for each stage should be chosen depending on each individual disease progression. A heuristic choice is that more weight should be given to $\beta_k$ if the stage features tend to be more discrete. 

\subsection{Non-regression disease stage sequence}\label{non-regression}
In the context of disease progression learning (Section \ref{disease_progression_learning}), we artificially define $S_0 > S_{K-1}$ to make the stage sequence cyclic, so that more variations of stage sequences can be generated for a particular disease to facilitate the neural network training. However, it is a bit counter-intuitive at the first thought to define $S_0 > S_{K-1}$, which violates the concept of monotonic disease progression. Alternatively, we could also use other approaches that do not require any similar assumptions, such as non-regression disease stage sequence, where the sequence can still start with any arbitrary disease stages and the disease simply stops progression when it reaches its final stage. In this way, the order of disease stages is still maintained without the assumption defined in the cyclic strategy. For example, $[ S_1, S_2, ..., S_{K-2}, S_{K-1}, S_{K-1} ]$ is also a valid non-regression stage sequence.
Therefore, a training sample of non-regression disease stage sequence can be formulated as the following:
\begin{equation*}\tag{9}\label{eq_new_sample_xy}
\begin{split}
\left( \boldsymbol{x}, \boldsymbol{y} \right) = \left( \right. [ I^{S_{\min(i+0, K-1)}}, I^{S_{\min(i+1, K-1)}}, ..., I^{S_{\min(i+K-1, K-1)}} ], \\ [ S_{\min(i+0, K-1)}, S_{\min(i+1, K-1)}, ..., S_{\min(i+K-1, K-1)} ] \left. \right)
\end{split}
\end{equation*}
where $i \in \left[ 0,K-1 \right]$ is the step shift size, indicating the starting stage of a sequence, and $K$ is the total number of disease stages. Although a non-regression disease stage sequence is more intuitive as compared to a cyclic stage sequence, we will show in Section \ref{results} that it does not outperform the cyclic stage sequence.

\subsection{Testing phase}
In the testing phase, given an image $I_{test}$, an artificial image sequence is generated by repeating $I_{test}$ for $K$ iterations, and then the sequence is fed into the trained network. Due to the design of our network, we have two options to predict its stage label based on either LSTM output $\hat{\boldsymbol{y}_l}$ or vision output $\hat{\boldsymbol{y}_v}$ (see Figure \ref{fig_architecture}). In both cases, only the first stage output in the sequence is reported as the final predicted label. Note that the input image sequence can be started with any arbitrary stage as we described earlier in Section \ref{disease_progression_learning} and Section \ref{non-regression}. Therefore a faked sequence starting with $I_{test}$ can still result in a reasonable prediction of stage label for $I_{test}$ based on its corresponding first stage output, although the rest stage outputs are invalid since there is no disease progression in the input sequence.

\subsection{Baseline network}
The baseline network for this study is the same vision model followed by the same softmax classifier $\mathcal{F}_v$ as used in our proposed network, except that the input is a single image, similar to most previous research work on medical image classification. 

For a fair comparison, we also make sure that both networks are trained on exactly the same amount of augmented data, which we will describe in detail in Section \ref{implementation_details}, to rule out the possibility that a superior performance could be gained simply due to larger training samples, given the fact that the input size of our proposed network is $K$ times bigger than that of the baseline network.

\section{Experiments and Results}
\subsection{Dataset}
To evaluate our proposed network, we choose to use a recently published dataset on diabetic retinopathy \cite{takahashi2017applying}, which contains fundus photographs of four stages. The dataset has two sets of class labels based on whether to grade with wider retinal area: Davis grading of one figure and Davis grading of concatenated figures, and we use the former in our experiments. There are in total 9939 images, with 6561 of NDR, 2113 of SDR, 460 of PPDR and 805 of PDR. 

Unlike previous work \cite{takahashi2017applying}, we address the data imbalance problem by undersampling by a random selection of 460 images from each stage class for each independent experiment, and moreover, all the images are resized to $200 \times 200$ instead of $1272 \times 1272$ to speed up training, since it is not our purpose here to compete with the result in \cite{takahashi2017applying}, but rather to investigate the advantage of disease progression learning in stage classification.

Out of the final resulting dataset, 10$\%$ of the data is randomly reserved for testing, and the rest is further split into two parts: 90$\%$ for training and 10$\%$ for validation. Data augmentation is performed on the training set by only using random rotations from $-$5$^\circ$ to 5$^\circ$ without shifts or flips, as all the images share the same position.

\subsection{Implementation details} \label{implementation_details}
We examine two types of vision models in our experiments: GoogleNet and ResNet-50, both of which were pretrained on ImageNet. The freeze layer is set to 64 for GoogleNet and 36 for ResNet-50. We use $C=256$ for the dimension of extracted feature vectors, $G=256$ for the LSTM model and $\boldsymbol{\alpha} = \boldsymbol{\beta} = \boldsymbol{1} \in \mathbb{R}^{K}$. All models are implemented in Keras with Theano backend, and trained using stochastic gradient descent, with the initial learning rate set to 0.001, decay by 1e-6 over each update, and Nestrov momentum is set to 0.9.

For each independent experiment, we evaluate the proposed network and baseline network on the same random split of the dataset. We repeat the above comparison experiment for 20 times, each with a different split. To make sure both networks are trained on the same amount of augmented data, the same number of iteration steps per epoch (i.e. 100 steps and 500 steps) is used in the training, until the validation loss does not drop for ten epochs (patience = 10, with the maximum number of epochs set to 100). In order to compensate the input size difference (i.e. one-image input against four-image input), the batch size for the baseline network training is set to 64 and decreased to 16 for our proposed network. Note that this is the only difference in hyper-parameter settings for the training of the two networks.

\subsection{Results}\label{results}
\begin{table*}[htbp]
	\caption{Performance comparisons of the baseline method and our proposed method trained on cyclic stage sequences (with both vision outputs and LSTM outputs).}
	\begin{center}
		\begin{tabular}{C{2.5cm} | C{1.5cm} | C{4.5cm} | C{4cm}}
			\hline
			Vision model & Steps $^{\mathrm{a}}$ & Method & \textit{Accuracy} \\
			\hline
			\multirow{6}{*}{\vtop{\hbox{\strut \text{ GoogleNet}}\hbox{\strut Inception v3}}} & \multirow{3}{*}{100} & \textsc{baseline} & $57.0 \pm 3.2 \%$ \\ \cline{3-4} 
			&                   & \textsc{Ours(vision output)} & $\textbf{59.4} \pm 3.1 \%$ \\ \cline{3-4} 
			&                   & \textsc{Ours(lstm output)} & $59.2 \pm 3.3 \%$ \\ \clineB{2-4}{3}
			& \multirow{3}{*}{500} & \textsc{baseline} & $59.5 \pm 3.3 \%$ \\ \cline{3-4} 
			&                   & \textsc{Ours(vision output)} & $\textbf{63.1} \pm 2.7 \%$ \\ \cline{3-4} 
			&                   & \textsc{Ours(lstm output)} & $62.9 \pm 3.0 \%$ \\ \clineB{1-4}{3}
			
			\multirow{6}{*}{ResNet-50} & 
			\multirow{3}{*}{100} & \textsc{baseline} & $57.3 \pm 6.4 \%$ \\ \cline{3-4} 
			&                   & \textsc{Ours(vision output)} & $\textbf{60.7} \pm 5.4 \%$ \\ \cline{3-4} 
			&                   & \textsc{Ours(lstm output)} & $60.7 \pm 5.7 \%$ \\ \clineB{2-4}{3} 
			& \multirow{3}{*}{500} & \textsc{baseline} & $58.1 \pm 4.8 \%$ \\ \cline{3-4} 
			&                   & \textsc{Ours(vision output)} & $61.4 \pm 3.8 \%$ \\ \cline{3-4} 
			&                   & \textsc{Ours(lstm output)} & $\textbf{61.7} \pm 3.9 \%$ \\ 
			\hline
			
			\multicolumn{4}{l}{$^{\mathrm{a}}$Training steps per epoch.} \\
		\end{tabular}
		\label{tab_accuracies}
	\end{center}
\end{table*}

Table \ref{tab_accuracies} shows the performance comparison of our proposed method and the baseline method, with both results of GoogleNet Inception v3 and ResNet-50 as the vision model. As shown in the table, our proposed method outperforms the baseline across all experimental settings that are used in this paper (choices of vision model and training steps per epoch), and on average, there is about 3.3$\%$ accuracy gain (2.4$\%$, 3.6$\%$, 3.4$\%$ and 3.6$\%$ for each setting respectively). However, we do not observe a significant performance difference between vision outputs and LSTM outputs (59.4$\%$ vs 59.2$\%$, 63.1$\%$ vs 62.9$\%$, 60.7$\%$ vs 60.7$\%$ and 61.4$\%$ vs 61.7$\%$) for our proposed method, probably due to the joint training of our model, so that the LSTM backpropagates the gradients to CNN to force it to learn features that are more stage-relevant, thus improving the vision outputs.

\begin{figure}[!b]
	\centering
	\includegraphics[scale=0.29]{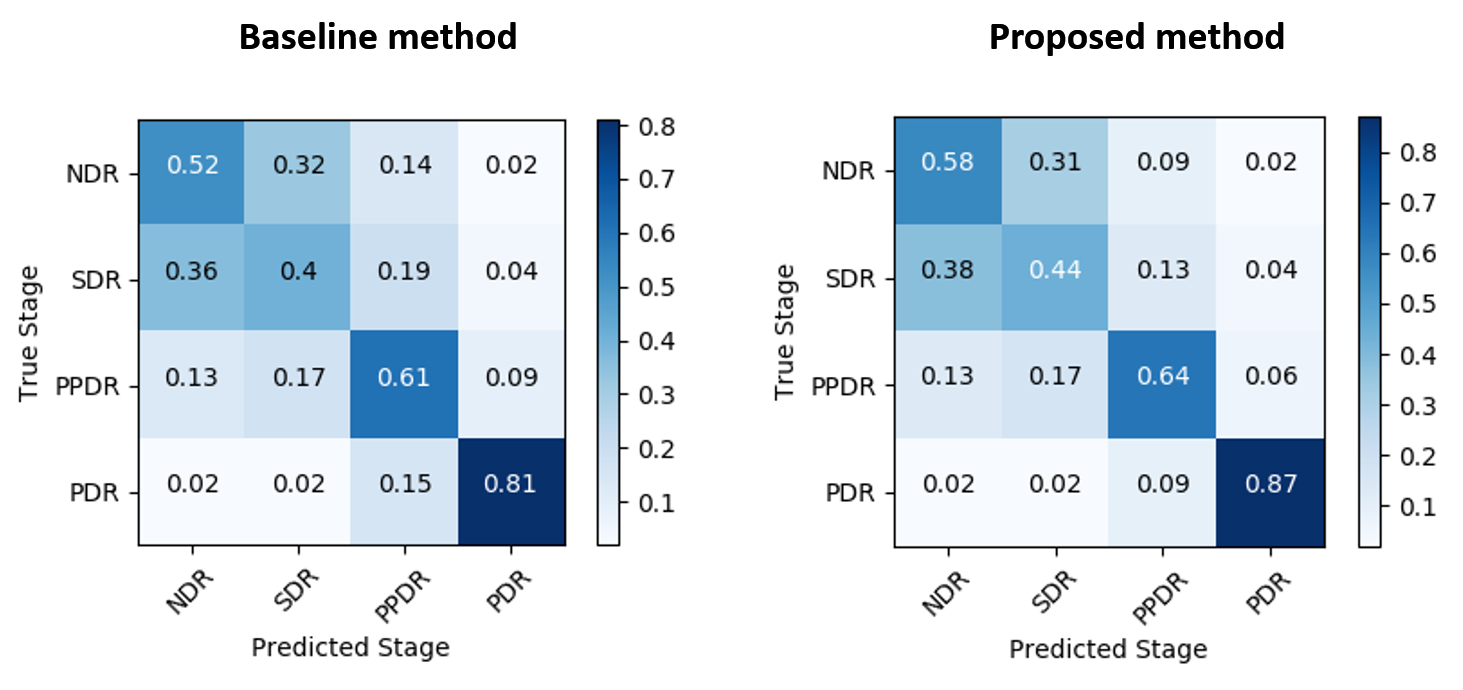}
	\caption{Confusion matrices of the baseline method (left) and our proposed method (right) using GoogleNet as the vision model with 500 training steps per epoch. }
	\label{fig_confusion_matrix}
\end{figure}
\begin{table*}[!b]
	\caption{Performance comparisons of the baseline method and our proposed method trained on non-regression stage sequences (with both vision outputs and LSTM outputs).}
	\begin{center}
		\begin{tabular}{C{2.5cm} | C{1.5cm} | C{4.5cm} | C{4cm}}
			\hline
			Vision model & Steps $^{\mathrm{a}}$ & Method & \textit{Accuracy} \\
			\hline
			\multirow{6}{*}{\vtop{\hbox{\strut \text{ GoogleNet}}\hbox{\strut Inception v3}}} & \multirow{3}{*}{100} & \textsc{baseline} & $\textbf{57.5} \pm 2.9 \%$ \\ \cline{3-4} 
			&                   & \textsc{Ours(vision output)} & ${56.2} \pm 3.8 \%$ \\ \cline{3-4} 
			&                   & \textsc{Ours(lstm output)} & $55.7 \pm 4.4 \%$ \\ \clineB{2-4}{3}
			& \multirow{3}{*}{500} & \textsc{baseline} & $59.1 \pm 3.5 \%$ \\ \cline{3-4} 
			&                   & \textsc{Ours(vision output)} & $\textbf{61.8} \pm 2.7 \%$ \\ \cline{3-4} 
			&                   & \textsc{Ours(lstm output)} & $60.9 \pm 2.5 \%$ \\ \clineB{1-4}{3}
			
			\multirow{6}{*}{ResNet-50} & 
			\multirow{3}{*}{100} & \textsc{baseline} & $56.5 \pm 5.8 \%$ \\ \cline{3-4} 
			&                   & \textsc{Ours(vision output)} & ${58.1} \pm 5.0 \%$ \\ \cline{3-4} 
			&                   & \textsc{Ours(lstm output)} & $\textbf{59.6} \pm 4.3 \%$ \\ \clineB{2-4}{3} 
			& \multirow{3}{*}{500} & \textsc{baseline} & $58.2 \pm 3.5 \%$ \\ \cline{3-4} 
			&                   & \textsc{Ours(vision output)} & $\textbf{60.5} \pm 2.5 \%$ \\ \cline{3-4} 
			&                   & \textsc{Ours(lstm output)} & ${60.1} \pm 2.8 \%$ \\ 
			\hline
			
			\multicolumn{4}{l}{$^{\mathrm{a}}$Training steps per epoch.} \\
		\end{tabular}
		\label{tab_accuracies2}
	\end{center}
\end{table*}

For both methods, accuracy performance can be improved with an increased number of training steps per epoch, which is within our expectation since more steps means more data augmentation. The best performance in our performed experiments is 63.1$\%$ for our proposed method and 59.5$\%$ for the baseline method, both of which are using GoogleNet Inception v3 pretrained on ImageNet with 500 training steps per epoch, and the confusion matrices are given in Figure \ref{fig_confusion_matrix}. It is also noted that most of the misclassified samples are located in NDR and SDR, which is reasonable since the two stages are often quite indistinguishable from a clinical perspective.

We believe that there is still room for performance improvement on this particular diabetic retinopathy problem by doing more data augmentation, using oversampling instead of undersampling (i.e. to make full use of images in the dataset), and using the original high resolution images. However, the purpose of this paper is to present and validate the idea of disease progression learning, and we leave the optimization for our next study.

To further test the alternative design of disease progression learning using non-regression stage sequences, we perform the comparison experiments using the same above settings, except that the model is trained on non-regression stage sequences (described in Section \ref{non-regression}), instead of cyclic stage sequences (described in Section \ref{disease_progression_learning}). As shown in Table \ref{tab_accuracies2}, the non-regression stage sequence fails to outperform cyclic stage sequence in all experiment settings, and in some case (GoogleNet and 100 training steps per epoch), it is even worse compared to the baseline method. The best result achieved by non-regression stage sequence is 61.8$\%$, which is still worse than that of cyclic sequence 63.1$\%$.

Based on the above results, we argue that despite being counter-intuitive at first thought, the cyclic strategy is merely to facilitate the training, which allows for cyclic variations of the training data that then de-emphasizes the first class as the first node, etc. This in turn allows the same disease progression sequence to be used for more training data. Since the change will be very abrupt from the last node to the first node, the learning will still retain the last node's characteristics as being substantially distinct from the first node of the sequence.

\balance
\section{Conclusion}
In this paper, we present a novel method for medical image recognition by leveraging disease progression learning, where stage sequences are learned by LSTM after feature extraction with a shared vision model for the images from each stage. Compared to the baseline method that is a pure vision model, our proposed method has an average of 3.3$\%$ accuracy increase based on our performed experiments, when evaluated for the problem of disease staging on a diabetic retinopathy dataset.

\bibliographystyle{splncs04} 
\bibliography{conference_041818}

\end{document}